\documentclass[letterpaper]{article} 
\usepackage[draft]{aaai25}  
\usepackage{times}  
\usepackage{helvet}  
\usepackage{courier}  
\usepackage[hyphens]{url}  
\usepackage{graphicx} 
\def\UrlFont{\rm}  
\usepackage{natbib}  
\usepackage{caption} 
\usepackage{algorithm}
\usepackage{amsmath}
\usepackage{algorithmic}
\usepackage{booktabs}
\usepackage{amsfonts}
\usepackage{multirow}
\usepackage[flushleft]{threeparttable}
\usepackage[table,xcdraw]{xcolor}
\usepackage{color}
\usepackage{tcolorbox}
\usepackage{colortbl}
\usepackage{dsfont}

\definecolor{cyan}{cmyk}{.3,0,0,0}
\newcommand{\VarSty}[1]{\textnormal{\ttfamily\color{cyan!90!black}#1}\unskip}
\usepackage{newfloat}
\usepackage{listings}

\DeclareCaptionStyle{ruled}{labelfont=normalfont,labelsep=colon,strut=off} 
\floatstyle{ruled}
\newfloat{listing}{tb}{lst}{}
\floatname{listing}{Listing}
\title{MTL-LoRA: Low-Rank Adaptation for Multi-Task Learning}
\author{
    Yaming Yang\textsuperscript{\rm 1}\equalcontrib \thanks{State Key Laboratory of General Artificial Intelligence, School of Intelligence Science and Technology.},
    Dilxat Muhtar\textsuperscript{\rm 2}\equalcontrib \thanks{Work done during internship at Microsoft.},
    Yelong Shen\textsuperscript{\rm 3}\equalcontrib,
    Yuefeng Zhan\textsuperscript{\rm 3},
    Jianfeng Liu\textsuperscript{\rm 3},
    Yujing Wang\textsuperscript{\rm 3} \thanks{Corresponding author.},
    Hao Sun\textsuperscript{\rm 3},
    Weiwei Deng\textsuperscript{\rm 3},
    Feng Sun\textsuperscript{\rm 3},
    Qi Zhang\textsuperscript{\rm 3},
    Weizhu Chen\textsuperscript{\rm 3},
    Yunhai Tong\textsuperscript{\rm 1}
}
\affiliations{
    \textsuperscript{\rm 1}Peking University \\
    \textsuperscript{\rm 2}Nanjing University \\
    \textsuperscript{\rm 3}Microsoft Corporation\\
    \textrm{yamingyang@stu.pku.edu.cn}, 
    \textrm{dmuhtar@smail.nju.edu.cn}, \\
    \textrm{\{yeshe, yuefzh, jianfengliu, yujwang\}@microsoft.com} \\ \textrm{yhtong@pku.edu.cn}

}

\begin{document}
\maketitle

\begin{abstract}
Parameter-efficient fine-tuning (PEFT) has been widely employed for domain adaptation, with LoRA being one of the most prominent methods due to its simplicity and effectiveness. 
However, in multi-task learning (MTL) scenarios, LoRA tends to obscure the distinction between tasks by projecting sparse high-dimensional features from different tasks into the same dense low-dimensional intrinsic space. 
This leads to task interference and suboptimal performance for LoRA and its variants.
To tackle this challenge, we propose MTL-LoRA, which retains the advantages of low-rank adaptation while significantly enhancing MTL capabilities. 
MTL-LoRA augments LoRA by incorporating additional task-adaptive parameters that differentiate task-specific information and capture shared knowledge across various tasks within low-dimensional spaces. This approach enables pre-trained models to jointly adapt to different target domains with a limited number of trainable parameters.
Comprehensive experimental results, including evaluations on public academic benchmarks for natural language understanding, commonsense reasoning, and image-text understanding, as well as real-world industrial text Ads relevance datasets, demonstrate that MTL-LoRA outperforms LoRA and its various variants with comparable or even fewer learnable parameters in MTL setting.

\end{abstract}

\begin{links}
    \link{Code}{https://github.com/pUmpKin-Co/MTL-LoRA}
\end{links}
\section{Introduction}

\begin{figure}[t!]
  \centering
  \includegraphics[width=\linewidth]{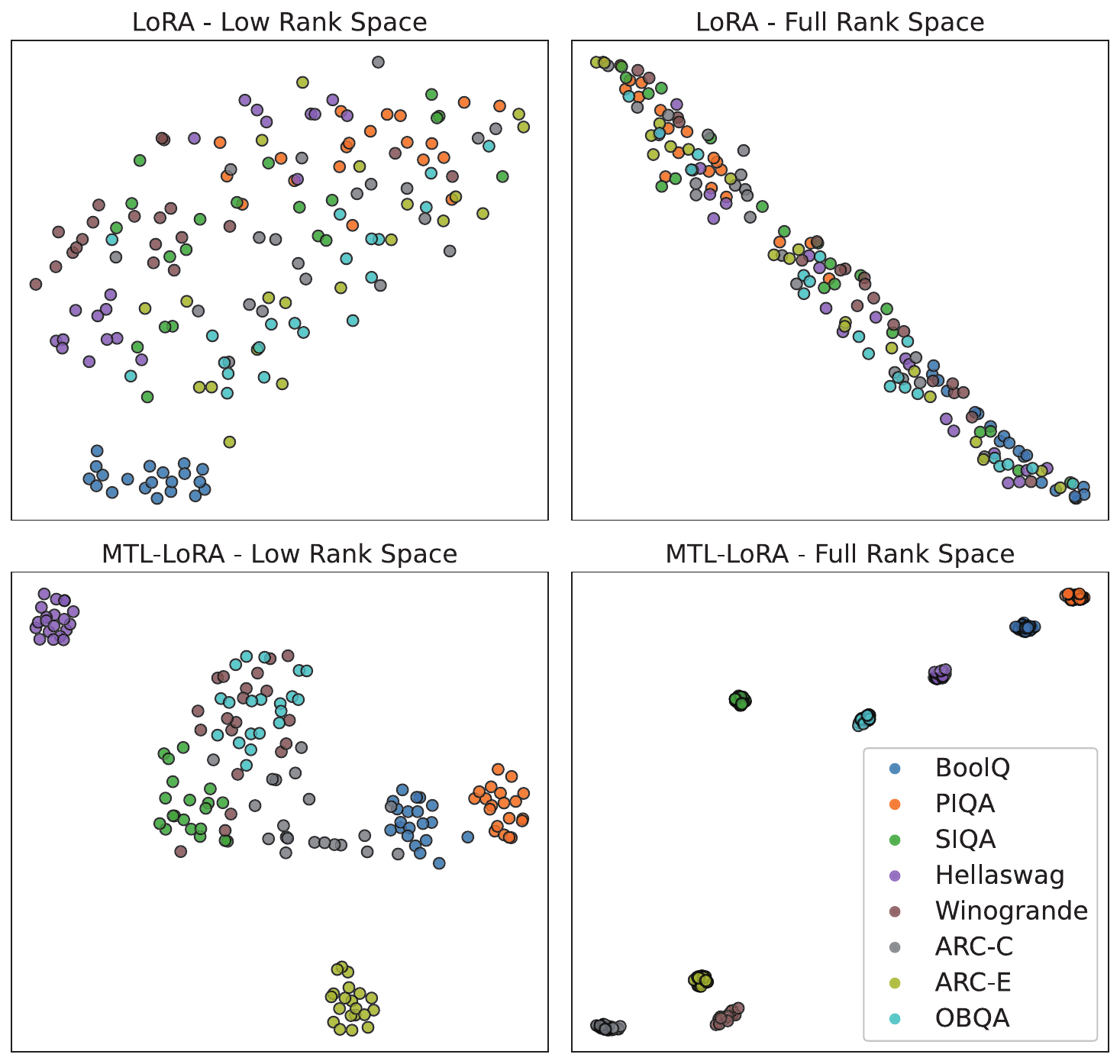}
  \caption{t-SNE visualization of task-specific features extracted from the $\mathbf{O}$ linear layer of the final block in the LLaMA2-7B model, comparing LoRA and MTL-LoRA after fine-tuning on a commonsense reasoning dataset.}
  \label{fig:leading}
  \vspace{-3mm}
\end{figure}

Large language models (LLMs)~\cite{brown2020language,touvron2023llama,MosaicML2023Introducing,Yang2024Qwen2TR,Abdin2024Phi3TR} now drive critical advances in artificial intelligence.
By scaling model size and leveraging extensive datasets, LLMs demonstrate exceptional generalization and advanced multi-task capabilities~\cite{wei2022emergent,Hoffmann2022TrainingCL}. 
The concept of “serving one model for different tasks” has led to numerous applications, ranging from natural language processing~\cite{Qin2023IsCA} to various domain-specific implementations~\cite{zhao2023survey,wei2022chain,min2023recent}.
Despite their high generalizability, LLMs still require fine-tuning for specific domains or to update their knowledge base. 
However, the vast number of parameters in LLMs poses significant challenges regarding computational efficiency and memory consumption during fine-tuning.

Parameter-efficient fine-tuning (PEFT) addresses this challenge by keeping the pre-trained model frozen and fine-tuning lightweight adapters~\cite{he2021towards,houlsby2019parameter}. A prominent PEFT approach is low-rank adaptation (LoRA)~\cite{hu2021lora}, which trains low-rank "adapter" layers in selected model components.
LoRA builds on the insight that fine-tuning updates in pre-trained LLMs exhibit low "intrinsic rank" during task specialization~\cite{aghajanyan2020intrinsic}, enabling effective approximation through targeted adapters. While LoRA and its recent variants~\cite{liu2024dora,Shi2024ResLoRAIR,Hayou2024LoRAEL} have shown promise across diverse LLM adaptation scenarios~\cite{Liu2023VisualIT,Huang2023LoraHubEC}, the growing task complexity and individual fine-tuning costs have spurred research into simultaneous multi-task LoRA training.
In this scenario, LoRA projects features from different tasks from a sparse high-dimensional space into a shared, dense low-dimensional space.
This projection causes interference and task confusion, amplifying the loss of task-specific information (as shown in the first row of Figure~\ref{fig:leading}). While information sharing is essential in multi-task learning (MTL), it is crucial that different task combinations exchange information distinctly. This necessitates additional design to enable LLMs to adaptively learn varied task information sharing strategies during fine-tuning.
Recent LoRA variants have attempted improvements in the multi-task setting, such as ensembling multiple LoRA adapters~\cite{wang2023multilora} or adopting Mixture of Experts (MoE) structures for soft information specification~\cite{Liu2023WhenMM}. However, these approaches do not effectively segregate task-specific information or implement distinct information sharing strategies, making them less effective in multi-task scenarios.

In this work, we introduce MTL-LoRA, a LoRA method designed to enhance LLMs with the capability to tackle a variety of tasks in a parameter-efficient manner. 
MTL-LoRA innovates by implementing task-specific transformations in low-rank space along with a strategy for adaptively exploring multiple information sharing methods. Specifically, MTL-LoRA begins by projecting inputs into lower intrinsic dimensions similar to LoRA. To mitigate the risk of cross-task information interference~\cite{hofmann2008kernel} within such a condensed space, MTL-LoRA introduces a learnable transformation for each task, ensuring the preservation of task-specific information.
Furthermore, acknowledging the role of information sharing among different tasks, especially in enhancing the performance of tasks with limited resources~\cite{crawshaw2020multi}, MTL-LoRA adopts a dynamic approach to learn different strategies for information sharing. 
With these improvements, MTL-LoRA efficiently assimilates both task-specific and shared information with minimal trainable parameters.
Comprehensive experiments on public academic benchmarks as well as real-world applications demonstrate that MTL-LoRA unleashes the multi-tasking capabilities of LLMs by fine-tuning a limited number of parameters, outperforming LoRA and its variants including.

The key contributions of our work can be summarized as follows:

\begin{itemize}
    \item[1] We present MTL-LoRA, which improves the capability of LoRA in MTL through an innovative approach for extracting task-specific information and enhancing cross-task information sharing.
    \item[2] Comprehensive experimental evaluations validate the efficacy of MTL-LoRA on both public academic benchmarks and real-world applications.
    \item[3] Through extensive ablation experiments and analyses, we confirm the effectiveness of each component of MTL-LoRA and validate the underlying design motivations.
\end{itemize}

\label{sec:intro}
\section{Related Work}
\subsection{Parameter-Efficient Fine-Tuning}
Parameter-efficient fine-tuning (PEFT) adapts large pre-trained models for specific tasks or domains using a small portion of parameters while keeping the main model frozen~\cite{he2021towards}.
A popular approach involves inserting trainable, continuous prompts or embeddings into the original text sequence to leverage the base model’s knowledge for new tasks~\cite{li2021prefix,liu2022p}. 
Another approach adds additional neural modules, like adapter structures, into pre-trained models~\cite{houlsby2019parameter,pfeiffer2020adapterfusion,lin2020exploring}.
In this trend, LoRA~\cite{hu2021lora} utilizes the concept of low intrinsic dimension~\cite{aghajanyan2020intrinsic}, introducing two trainable rank decomposition matrices into frozen pre-trained models to estimate the accumulated gradient update during fine-tuning.
Due to its lower inference latency and superior performance, LoRA has been widely adopted, and many studies are exploring ways to enhance its efficiency and stability.
For example, AdaLoRA~\cite{Zhang2023AdaLoRAAB} incorporates an importance-aware rank allocation method to assign ranks according to layer importance. LoRA+~\cite{Hayou2024LoRAEL} aims to improve LoRA’s training stability by using different learning rates for different low-rank matrices. DoRA~\cite{liu2024dora} further enhances both the learning capacity and training stability of LoRA by decomposing the pre-trained weights into two components, magnitude and direction, for fine-tuning.
While LoRA and its variants show promise in single-task adaptations, their effectiveness diminishes in multi-task scenarios as they update parameters uniformly across all tasks, overlooking crucial task-specific information and dynamic task information sharing. Our work focuses on improving LoRA to acquire both task-specific and task-agnostic knowledge, enhancing its performance in multi-task settings.

\subsection{Multi-Task Learning}
Multi-Task Learning (MTL) aims to optimize all tasks jointly and adapt a single trained model to serve for all tasks~\cite{crawshaw2020multi}.
Since language models trained on large-scale datasets can extract universal representations, previous multi-task learning methods, such as MT-DNN~\cite{liu-etal-2019-multi}, typically use a shared pre-trained model with task-specific heads to jointly adapt the model to different tasks.
However, as the size of pre-trained models continues to increase, full fine-tuning (FT) introduces significant computational overhead and an increased risk of catastrophic forgetting~\cite{Biderman2024LoRALL}. 
While LoRA offers an alternative to FT, it does not perform as well in multi-task settings~\cite{wang2023multilora}. 
Approaches like MultiLoRA~\cite{wang2023multilora} and MoELoRA~\cite{Liu2023WhenMM} improve LoRA’s multi-task performance in joint training scenarios by integrating multiple LoRAs or utilizing expert routing. However, they fail to strike a good balance between task-specific information and task-information sharing, resulting in suboptimal performance.

\label{sec:related_Work}
\section{Method}
In this section, we first introduce the low-rank adaptation method, followed by an in-depth explanation of the proposed MTL-LoRA for multi-task learning.

\begin{figure}[t]
  \centering
  \includegraphics[width=\linewidth]{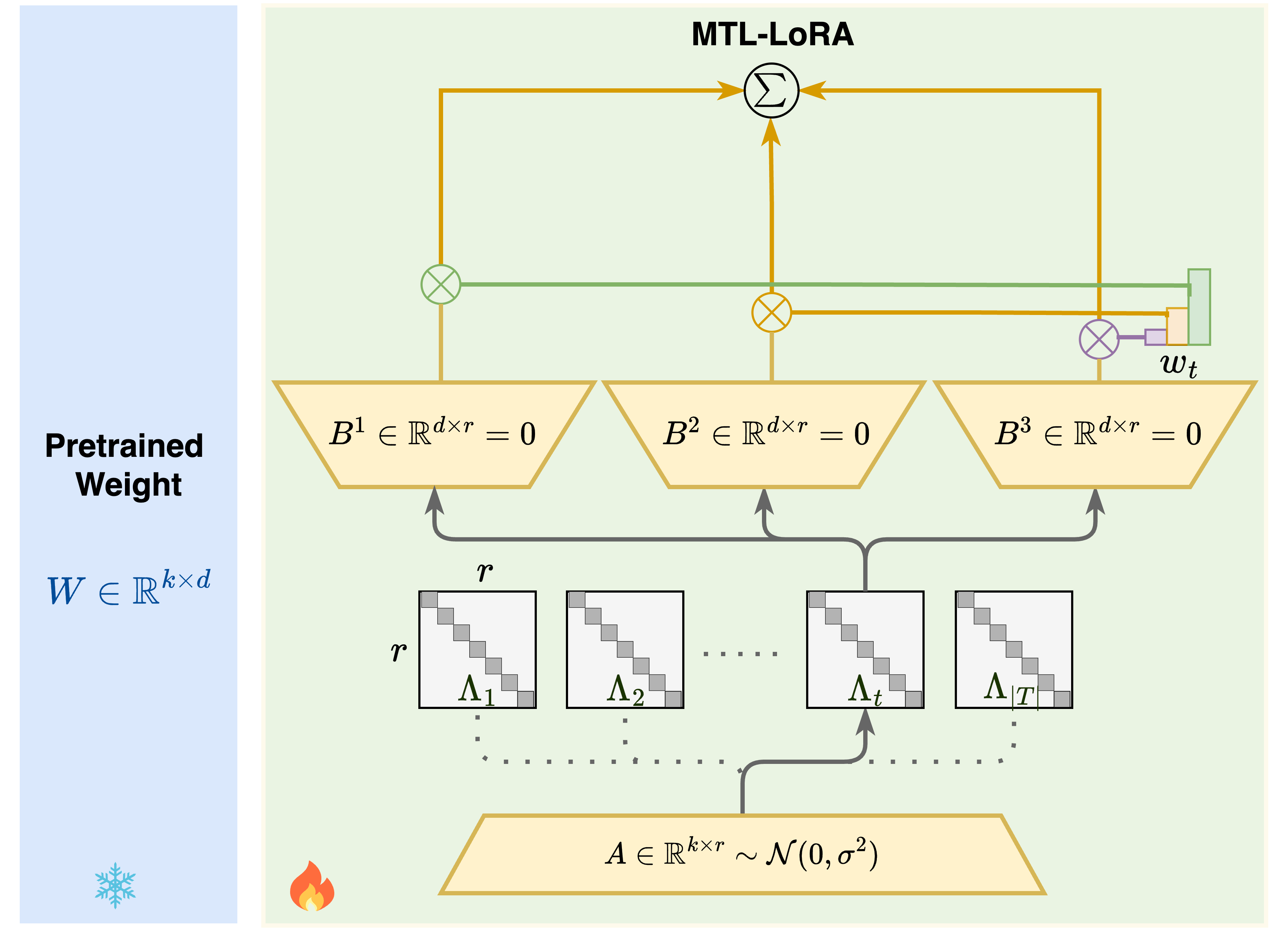}
 \caption{The overall architecture of MTL-LoRA. MTL-LoRA employs task-specific transformation matrices and multiple up-projection matrices to learn both task-specific and shared information.}
  \label{fig:architecture}
  \vspace{-3mm}
\end{figure}

\subsection{Low-Rank Adaption}
Current LLMs generally follow a decoder-only structure, characterized by a series of blocks, each comprising two key components with residual connections: a multi-head self-attention (MHA) layer and a feed-forward network (FFN)~\cite{brown2020language,touvron2023llama,MosaicML2023Introducing}. The MHA layer involves using dense learnable matrices $\mathbf{W}_q$, $\mathbf{W}_k$, $\mathbf{W}_v$, and $\mathbf{W}_o$ to mix the sequence $x$ according to inter-relationships between tokens:
\begin{align}
    \begin{split}
    \text{MHA}(\mathrm{x}) = \text{Softmax}\left( \frac{(\mathbf{W}_q\mathrm{x})^T\mathbf{W}_k\mathrm{x}}{\sqrt{k}}\right)(\mathbf{W}_v\mathrm{x})^T\mathbf{W}_o,
    \end{split}
\end{align}
where we assume a single attention head and $k$ denotes the hidden dimension for the head.
The FFN layer is usually an MLP with two dense linear projection layers, $\mathbf{W}_{down}$ and $\mathbf{W}_{up}$, and a non-linear activation function $\sigma(\cdot)$ for channel mixing:
\begin{align}
    \begin{split}
    \text{FFN}(\mathrm{x}) = \sigma(\mathrm{x}\mathbf{W}_{down})\mathbf{W}_{up}.
    \end{split}
\end{align}

Although LLMs pre-trained with extensive general domain datasets have demonstrated remarkable generalization abilities, there is a need to adapt these models for specific tasks or domains with limited resources. To achieve this, low-rank adaptation (LoRA), inspired by the concept of low intrinsic dimensionality in LLMs, decomposes the weight gradient $\Delta\mathbf{W}$ into low-rank matrices, thereby reducing the number of trainable parameters.
Specifically, for a dense weight matrix $\mathbf{W} \in \mathbb{R}^{d \times k}$, LoRA employs two low-rank matrices, $\mathbf{B} \in \mathbb{R}^{d \times r}$ and $\mathbf{A} \in \mathbb{R}^{r \times k}$, to approximate the accumulated gradient updates $\Delta\mathbf{W}$. The rank $r$ is chosen to be much smaller than the minimum of $d$ and $k$, effectively decreasing the number of trainable parameters. Consequently, the resulting weight matrix is expressed as $\mathbf{W} + \mathbf{B}\mathbf{A}$, and the output $h$ for an input $x$ through this updated weight matrix is formulated as:
\begin{equation}
    \begin{aligned}
        h &= (\mathbf{W} + \Delta\mathbf{W})x = \mathbf{W}x + \mathbf{BA}x
    \end{aligned}
\end{equation}
In implementation, the low-rank matrix $\mathbf{A}$ is initialized with Kaiming Uniform~\cite{he2015delving} and $\mathbf{B}$ is initialized with zero to ensure $\Delta\mathbf{W} = 0$ at the start, thereby contributing to training stability. 
A constant scale factor $ \alpha $ is also introduced to adjust the magnitude of the changes of the updated matrix $\Delta\mathbf{W} $ made by LoRA modules.

\begin{table*}[h!]
\centering
\resizebox{0.9\textwidth}{!}{%
\begin{tabular}{@{}lcccccccccc@{}}
\toprule
Model   \& Method & \# Trainable (\%)     & CoLA            & MNLI            & MRPC            & QNLI            & QQP             & RTE             & SST2            & STSB            & \multirow{2}{*}{Avg.} \\ \cmidrule(r){1-10}
                  &        & Mcc.            & Acc             & Acc             & Acc             & Acc             & Acc             & Acc             & Pea.            &                       \\ \midrule
MPT-FT            & 100 $\times$ 8   & 0.6712          & 0.9105          & 0.9044          & 0.9551          & 0.9213          & 0.9036          & 0.9701          & 0.9148          & 0.8939                \\ 
MPT-LoRA-ST          & 0.12 $\times$ 8 & 0.6161          & 0.9097          & 0.8578          & 0.9588 & 0.9033          & 0.8881          & 0.9636          & 0.8858          & 0.8729                \\ \midrule
MPT-LoRA-MT         & 0.24           & 0.6528          & 0.9102          & 0.8775          & \textbf{0.9590}          & 0.9094          & \textbf{0.9206}          & \textbf{0.9679} & 0.9173          & 0.8893                \\
MPT-MultiLoRA     & 0.38           & 0.6441          & 0.9092          & 0.8701          & 0.9564          & 0.9085          & 0.9061          & \textbf{0.9679} & 0.9148          & 0.8846                \\ 
MPT-MoELoRA    & 0.24           & 0.6421          & 0.9096          & 0.8695         & 0.9570          & 0.9119          & 0.9159          & 0.9639         & 0.8872          & 0.8820
\\
MPT-MTL-LoRA     &  0.24           & \textbf{0.6754} & \textbf{0.9104} & \textbf{0.8995} & 0.9584          & \textbf{0.9122} & 0.9170          & 0.9644          & \textbf{0.9230} & \textbf{0.8950}       \\ \midrule \midrule
LLaMA2-FT         & 100 $\times$ 8    & 0.7009          & 0.9144          & 0.8480          & 0.9690          & 0.9273          & 0.8773          & 0.9701          & 0.8965          & 0.8879                \\ 
LLaMA2-LoRA-ST       & 0.12 $\times$ 8 & 0.6266          & 0.9099          & 0.8484          & 0.9518          & 0.9049          & 0.8989          & 0.9667          & 0.8915          & 0.8748                \\ \midrule
LLaMA2-LoRA-MT      & 0.24           & 0.6591          & 0.9139          & 0.8603          & 0.9600          & 0.9088          & 0.9170          & 0.9713          & 0.9188          & 0.8887                \\
LLaMA2-MultiLoRA  & 0.38           & 0.6134          & 0.9100          & 0.8628          & 0.9552          & 0.9003          & 0.9097          & 0.9633          & 0.9184          & 0.8791                \\ 
LLaMA2-MoELoRA   & 0.24  & 0.6366        & 0.9118          & 0.8554          & 0.9568      & 0.9062 & 0.9206 & 0.9656   & 0.9218 & 0.8844 
\\
LLaMA2-MTL-LoRA  &  0.24          & \textbf{0.6797} & \textbf{0.9143} & \textbf{0.9020} & \textbf{0.9628} & \textbf{0.9142} & \textbf{0.9242} & \textbf{0.9713} & \textbf{0.9279} & \textbf{0.8996}       \\ \bottomrule
\end{tabular}%
}
\caption{Performance of different adaption methods on the GLUE benchmark, with results reported for the validation set. FT: full parameters fine-tuning. ST: single task fine-tuning. MT: multi-task fine-tuning.
}
\label{tab:com-glue}
\end{table*}

\begin{table*}[t!]
\centering
\resizebox{0.9\textwidth}{!}{%
\begin{tabular}{@{}lcccccccccc@{}}
\toprule
 Method    & \# Trainable(\%) & BoolQ         & PIQA          & SIQA          & Winogrande    & OBQA          & Hellaswag     & ARC-E         & ARC-C         & Avg.          \\ \midrule
                                                LoRA$^{\dagger}$      & 0.83             & 69.8 &   79.9        & 79.5          & 82.6          & 81.0          & 83.6          & 79.8          & 64.7          & 77.6          \\
                                                                           DoRA$^{\dagger}$      & 0.43             & \textbf{72.0}          & 83.1          & 79.9          & 83.0          & 81.2          & 89.1          & 84.5          & 71.0          & 80.5          \\
                                                                           MultiLoRA & 0.40             & 66.5          & 65.8          & 62.8          & 79.3          & 75.4          & 79.2          & 76.7          & 59.6          & 70.7          \\
                                                                           MoELoRA   & 0.25             & 68.0          & 83.5          & 70.4          & 82.5          & \textbf{83.2}          & 90.6          & 86.8          & 61.5          & 78.3          \\ 
                                                                        MTL-LoRA  & 0.25             & 71.0          & \textbf{84.4} & \textbf{80.8} & \textbf{84.9} & 82.6          & \textbf{93.1} & \textbf{87.0} & \textbf{73.4} & \textbf{82.1} \\ \bottomrule
\end{tabular}%
}
\caption{Commonsense reasoning results. We follow the setting from~\cite{liu2024dora,hu2023llm} for jointly training all tasks.$^{\dagger}$ means the results from original DoRA paper~\cite{liu2024dora}.}
\label{tab:commonsense}
\end{table*}

\subsection{MTL-LoRA}
While LoRA effectively fine-tunes LLMs for specific domains using minimal trainable parameters, it does not fully accommodate the dynamics of task-specific and shared knowledge within different tasks or domains, thereby limiting its effectiveness in MTL settings. 
To address this issue, we introduce MTL-LoRA to improve LoRA with enhanced MTL abilities. 
The architecture of the proposed MTL-LoRA is detailed in Figure \ref{fig:architecture}.
For a given input $\mathrm{x}_t $ corresponding to task $t$,  
MTL-LoRA projects $\mathrm{x}_t$ to low dimension space through $\mathbf{A}$ as in LoRA.
However, to enhance the differentiation of tasks within this low, information-dense feature space and maintain task-specific information, MTL-LoRA incorporates a low-rank learnable matrix $\Lambda_t \in \mathbb{R}^{r \times r}$ for each task. 
This process involves transforming the projected sample $\mathbf{A}\mathrm{x}_t$ via $\Lambda_t$ to isolate information pertinent to the specific task. 
Furthermore, we argue that diverse information-sharing strategies are critical for leveraging knowledge from different tasks to improve overall performance.
Therefore, rather than relying on a single up-project matrix for information aggregation, we utilize multiple low-rank matrices to explore various information sharing strategies.
These combinations are then integrated using a weighted averaging strategy, thereby facilitating adaptive information sharing among different tasks.
Specifically, assuming that the up-projection matrix is denoted as $\mathbf{B}^i \in \mathbb{R}^{d \times r}$ and the learnable averaging weight for task $t$ is represented by $w_t \in \mathbb{R}^{n \times 1}$, where $n$ is the number of up-projection low-rank matrices, the output of MTL-LoRA for task $t$ is formulated as:
\begin{equation}\label{eq:mtl}
    \begin{aligned}
        h_t &= (\mathbf{W} + \Delta\mathbf{W}_t)x_t \\
          &= \mathbf{W}x_t + \sum_{i=1}^n\frac{\text{exp}(w_t^i/\tau)\mathbf{B}^i}{\sum_{j=1}^n\text{exp}(w_t^{j}/\tau)}\Lambda_t\mathbf{A}x_t
    \end{aligned}
\end{equation}
where $\tau$ is a hyperparameter to control the sharpness of the weight distribution, and the superscript represents the indices of the corresponding up-projection matrix and averaging weight. We initialize $\Lambda_i$ as a diagonal matrix with each diagonal element being 1, thereby ensuring that $\Delta\mathbf{W} = 0$ at the start of training.
Building upon these advancements, MTL-LoRA maintains the benefits of parameter efficiency while substantially boosting the MTL capabilities of LoRA.\label{sec:method}

\section{Experiments}
We conduct a series of experiments to demonstrate the effectiveness of MTL-LoRA on various tasks, including natural language understanding (NLU), commonsense reasoning, and image-text understanding. Additionally, we perform ablation studies to illustrate the effectiveness of each component of MTL-LoRA. Finally, we conduct a sensitivity analysis to examine its stability across different hyperparameter configurations.

\begin{table*}[ht!]
\centering
\resizebox{\textwidth}{!}{%
\begin{tabular}{@{}lcccccccccccccccc@{}}
\toprule
\multirow{3}{*}{Model \& Method}    & \multirow{3}{*}{\# Trainable (\%)} & \multicolumn{15}{c}{Task Index} \multirow{3}{*}{Avg.} \\ \cmidrule(r){3-16} & & 0               & 1               & 2               & 3               & 4               & 5               & 6               & 7               & 8               & 9               & 10              & 11              & 12              & 13              & \\ 
\cmidrule(r){3-16}
                      & \multicolumn{15}{c}{AUC-ROC}                                                                                                                                                                                                                              &                       \\ \midrule
MPT-LoRA-ST           & 0.12 $\times$ 14   & 0.8846          & 0.8361          & 0.8979          & 0.8868          & 0.8806          & 0.8833          & 0.8941          & 0.8589          & 0.8677          & 0.8676          & 0.8490          & 0.8699          & 0.8519          & 0.8689          & 0.8712                \\ \midrule
MPT-LoRA-MT          & 0.24       & 0.8812          & 0.8286          & 0.8863          & 0.8722          & 0.8765          & 0.8827          & 0.8858          & 0.8576          & 0.8584          & 0.8590          & 0.8399          & 0.8680          & 0.8504          & 0.8639          & 0.8650                \\
MPT-MultiLoRA          & 0.38       & 0.8874          & 0.8382          & 0.8947          & 0.8871          & 0.8835          & 0.8890          & 0.8944          & 0.8674          & 0.8680          & 0.8712          & 0.8538          & 0.8786          & 0.8598          & 0.8703          & 0.8745                \\ 
MPT-MoELoRA            & 0.24        & 0.8860          & 0.8382    & 0.8898     & 0.8837     & 0.8829 & 0.8889      &  0.8950    & 0.8641    & 0.8682     & 0.8720         & 0.8540      & 0.8802     & 0.8576    & \textbf{0.8769}     & 0.8741
\\
MPT-7B-MTL-LoRA & 0.24       & \textbf{0.8876} & \textbf{0.8384} & \textbf{0.9005}$^{*}$ & \textbf{0.8918}$^{*}$ & \textbf{0.8840}$^{*}$ & \textbf{0.8897}$^{*}$ & \textbf{0.8956}$^{*}$ & \textbf{0.8678} & \textbf{0.8692}$^{*}$ & \textbf{0.8728}$^{*}$ & \textbf{0.8561}$^{*}$ & \textbf{0.8810}$^{*}$ & \textbf{0.8607}$^{*}$ & 0.8714 & \textbf{0.8762}       \\ \midrule \midrule
LLaMA2-LoRA-ST        & 0.12 $\times$ 14   & 0.8838          & 0.8349          & 0.8992          & 0.8886          & 0.8792          & 0.884           & 0.8948          & 0.8575          & 0.8665          & 0.8685          & 0.8495          & 0.8706          & 0.8523          & 0.8703          & 0.8714                \\ \midrule
LLaMA2-LoRA-MT       & 0.24       & 0.8842          & 0.8329          & 0.8932          & 0.8867          & 0.8806          & 0.8859          & 0.8935          & 0.8646          & 0.8652          & 0.8671          & 0.8545          & 0.8763          & 0.8571          & 0.8689          & 0.8722                \\
LLaMA2-MultiLoRA          & 0.38       & 0.8879          & 0.8369          & 0.8958          & 0.8922          & 0.8853          & \textbf{0.8901} & 0.8950          & 0.8687          & 0.8693          & 0.8743          & 0.8540          & 0.8808          & 0.8599          & \textbf{0.8726} & 0.8759                \\ 
LLaMA2-MoELoRA    & 0.24        & 0.8850           & 0.8370 & 0.8962 & 0.8921 & 0.8850   & 0.8876      & 0.8961           & 0.8685 & 0.8686 & 0.8737 & 0.8550 & 0.8800   & 0.8594      & 0.8722           & 0.8755
\\
LLaMA2-MTL-LoRA   & 0.24       & \textbf{0.8883} & \textbf{0.8374} & \textbf{0.9016}$^{*}$ & \textbf{0.8929}$^{*}$ & \textbf{0.8856} & 0.8899          & \textbf{0.8968}$^{*}$ & \textbf{0.8688} & \textbf{0.8706}$^{*}$ & \textbf{0.8754}$^{*}$ & \textbf{0.8582}$^{*}$ & \textbf{0.8810} & \textbf{0.8607}$^{*}$ & 0.8723          & \textbf{0.8771}       \\ \bottomrule 
\end{tabular}%
}
\caption{Results of different methods on the test set of Ads dataset. $^{*}$  denotes that the significance test is passed at a 90\% confidence level. ST: single task fine-tuning. MT: multi-task fine-tuning.}
\label{tab:com-ads}
\end{table*}

\begin{table}[ht!]
\centering
\resizebox{\columnwidth}{!}{%
\begin{tabular}{@{}lcccccc@{}}
\toprule
Method & \# Trainable (\%)     & VQA$^{\text{v2}}$            & GQA            & NVLR$^{\text{2}}$            & CoCo Cap       & Avg. \\
\midrule
FT$^{\dagger}$ & 100 & 66.9 & 56.7 & 73.7 & 112.0 & 77.3 \\
\midrule
LoRA$^{\dagger}$     & 5.93  &  65.2 & 53.6 & 71.9 & 115.3 & 76.5  \\ 
DoRA$^{\dagger}$     & 5.96 & 65.8 & 54.7 & \textbf{73.1} &\textbf{ 115.9} & 77.4 \\
MTL-LoRA     &  5.19  & \textbf{68.6} & \textbf{54.9} & 72.6 & 114.6 & \textbf{77.7}    \\ \midrule 
\end{tabular}%
}
\caption{The multi-task evaluation results on VQA, GQA, NVLR$^{2}$, and CoCo Caption using the VL-BART backbone. $^{\dagger}$ indicates results taken from the original DoRA paper~\cite{liu2024dora}.}
\label{tab:com-vl}
\vspace{-3mm}
\end{table}

\subsection{Evaluation on Public Benchmark}
\subsubsection{Natural Language Understanding}\label{subsec:nlu}
We compare MTL-LoRA against several baseline methods, including full-parameter tuning, single-task fine-tuning with LoRA, multi-task fine-tuning with LoRA, MultiLoRA, and MoELoRA, on both MPT-7B~\cite{MosaicML2023Introducing} and LLaMA2-7B~\cite{touvron2023llama} models.
We use the widely recognized GLUE~\cite{wang2018glue} benchmark for evaluation. The GLUE benchmark comprises nine NLU tasks, covering a diverse range of linguistic challenges such as sentiment analysis, textual entailment, and sentence similarity.

To ensure that the LLM with decoder-only architecture generates stable classification results, we adopt the approach of MT-DNN~\cite{liu-etal-2019-multi} by assigning each task its respective classification head. 
To harness the generative capabilities of LLMs, we reformat each task using a specific template and initialize the weights of the classification head with the corresponding word embeddings of the target answer from the original language model. 
For each PEFT method, we train only the adapter parameters and the task-specific classification head. 
Details on hyperparameters and templates can be found in Section A.1 and Section C of the supplementary material.

The results presented in Table~\ref{tab:com-glue} demonstrate that MTL-LoRA achieves superior performance, surpassing other baseline methods across both LLMs. Notably, MTL-LoRA not only outperforms the strong MultiLoRA baseline with only 64\% trainable parameters but also exceeds the performance of FT while requiring significantly fewer trainable parameters (merely 0.03\% per task compared to FT). Furthermore, as shown in Table~\ref{tab:com-glue}, MTL with LoRA (i.e., LoRA-MT) outperforms single task fine-tuning with LoRA (i.e., LoRA-ST) across all eight tasks and largely reduces the number of adapters that need to be maintained.

While both MoELoRA and MultiLoRA have boosted LoRA’s multi-tasking performance, optimizing these models for multi-task scenarios remains challenging, leading to suboptimal performance. In contrast, MTL-LoRA effectively exploits both task-specific and task-agnostic information, thereby outperforming LoRA-MT in nearly all tasks.

\subsubsection{Commonsense Reasoning}
In these experiments, we perform a comparison of MTL-LoRA against LoRA and various LoRA variants on LLaMA2-7B for commonsense reasoning tasks. We train each model on eight sub-tasks jointly and evaluate performance on the individual test dataset for each task.
Following the same train-test split protocol and instruction prompts as in~\cite{hu2023llm,liu2024dora}, we report the test set accuracy for each method. Detailed hyperparameter settings can be found in Section A.2 of the supplementary material. Where possible, we report model results as presented in the original papers.

The results in Table~\ref{tab:commonsense} show that both MoELoRA and DoRA outperform LoRA and MultiLoRA. We hypothesize that this is because commonsense reasoning tasks require fine-grained task routing or gradient decomposition, rather than merely ensembling different LoRAs. Despite this, MTL-LoRA consistently outperforms all baseline methods.
Notably, with only one-third of the LoRA parameters, MTL-LoRA surpasses LoRA by approximately 4\%. Moreover, with only half the parameters, MTL-LoRA outperforms the strong baseline DoRA by a large margin of 2\%. 
These results further highlight MTL-LoRA’s efficiency and effectiveness in optimizing model performance while minimizing training overhead in multi-task adaptation.

\subsubsection{Image-Text Understanding}
To evaluate the performance of MTL-LoRA in a multimodal multitask fine-tuning context, we compare it with LoRA, DoRA, and FT using VL-BART in four distinct image text tasks. The results for FT, LoRA, and DoRA are taken from the original DoRA paper. For MTL-LoRA, we follow the same settings as DoRA, applying the adapter to the $\mathbf{Q}$ and $\mathbf{V}$ linear layers of the language model. We also unfreeze the bias and layer normalization parameters, training for 20 epochs with the AdamW optimizer and a learning rate of $1\times 10^{-3}$, in line with DoRA’s configuration. The rank, alpha, number of up-projection matrices, and temperature for MTL-LoRA are set to 64, 16, 2, and 0.8, respectively. 

The results are presented in Table~\ref{tab:com-vl}. Both MTL-LoRA and DoRA outperform LoRA by 1\% with the same or even fewer parameters. Furthermore, MTL-LoRA and DoRA surpass FT while unfreezing only 5-6\% of the model’s parameters. In the multimodal multitasking scenario, MTL-LoRA also outperforms DoRA with approximately 1\% fewer learnable parameters, demonstrating its learning effectiveness in this setting.

\begin{table*}[t!]
\centering
\resizebox{0.85\textwidth}{!}{%
\begin{tabular}{lccccccccc}
\hline
Method  
& CoLA            & MNLI            & MRPC            & QNLI            & QQP             & RTE             & SST2            & STSB            & \multirow{2}{*}{Avg.} \\ \cline{2-9}                      & Mcc.            & Acc.            & Acc.            & Acc.            & Acc.            & Acc.            & Acc.            & Pea.            &                       \\ \hline
LoRA-MT  & 0.6591 & 0.9193 & 0.8603 & 0.9600 & 0.9088 & 0.9170 & 0.9713 & 0.9188  &0.8887            \\\hline
MTL-LoRA & \textbf{0.6797} & 0.9143          & \textbf{0.9020} & \textbf{0.9628} & \textbf{0.9142} & \textbf{0.9242} & \textbf{0.9713}          & \textbf{0.9279} & \textbf{0.8996}       \\
w/o $\Lambda_t$  & 0.6567          & \textbf{0.9164} & 0.8971          & 0.9627          & 0.9140          & \textbf{0.9242} & 0.9679          & 0.9261          & 0.8956   (\textcolor{red}{-0.0039})             \\
w/o $\tau$      & 0.6592          & 0.9144          & 0.8676          & 0.9606          & 0.9114          & 0.9097          & \textbf{0.9713}          & 0.9214          & 0.8895 (\textcolor{red}{-0.0101})               \\
$n=1$           & 0.6360          & 0.9150          & 0.8725          & 0.9625          & 0.9105          & 0.9170          & 0.9702          & 0.9248          & 0.8886 (\textcolor{red}{-0.0110})               \\ \hline
\end{tabular}%
}
\caption{Results of ablation studies on MTL-LoRA across from the GLUE benchmark. MT: multi-task fine-tuning.}
\label{tab:ablation}
\end{table*}

\begin{table*}[t!]
\centering
\resizebox{0.85\textwidth}{!}{%
\begin{tabular}{@{}lccccccccccccccc@{}}
\toprule
\multirow{3}{*}{Method \& Model} & \multicolumn{14}{c}{Task Index} & \multirow{3}{*}{Avg.} \\ \cmidrule(r){2-15}  &  0             & 1             & 2             & 3             & 4             & 5             & 6             & 7             & 8             & 9             & 10            & 11            & 12            & 13            &  \\ \cmidrule(r){2-15}
                 & \multicolumn{14}{c}{F1}                                                                                                                                                                                                       &                       \\ \midrule 
LLaMA2-LoRA-MT     & 0.60          & 0.17          & 0.58          & 0.37          & 0.38          & 0.52          & 0.59          & 0.18          & 0.61          & 0.71          & \textbf{0.44}          & 0.13          & 0.18          & 0.12          & 0.39                 \\
LLaMA2-MTL-LoRA & \textbf{0.86} & \textbf{0.84} & \textbf{0.81} & \textbf{0.99} & \textbf{0.99} & \textbf{0.97} & \textbf{0.94} & \textbf{0.66} & \textbf{0.99} & \textbf{0.93} & 0.43 & \textbf{0.73} & \textbf{0.34} & \textbf{0.55} & \textbf{0.79}         \\ \midrule \midrule
MPT-LoRA-MT        & 0.59          & 0.27          & 0.60          & 0.34          & 0.36          & 0.49          & 0.52          & 0.30          & 0.51          & 0.68          & 0.26          & 0.23          & 0.14          & 0.14          &  0.39                 \\
MPT-MTL-LoRA    & \textbf{0.83} & \textbf{0.92} & \textbf{0.81} & \textbf{0.98} & \textbf{0.97} & \textbf{0.86} & \textbf{0.65} & \textbf{0.75} & \textbf{0.87} & \textbf{0.83} & \textbf{0.39} & \textbf{0.75} & \textbf{0.72} & \textbf{0.45} & \textbf{0.77}  \\ \bottomrule
\end{tabular}%
}
\caption{Task classification results on Ads dataset using SVM classifier with features extracted by various methods. All results are averaged over five run with different train-test split seed. MT: multi-task fine-tuning.}
\label{tab:svm}
\end{table*}

\subsection{Evaluation on In-house Dataset}
To further validate the performance of MTL-LoRA in large-scale, complex multi-task scenarios, we compared different multi-task low-rank adaptation strategies on an in-house text Ads relevance dataset (referred to as the Ads dataset).

The text Ads relevance task involves determining whether a query is semantically relevant to a given Ad. This dataset encompasses 14 tasks, covering various production scenarios. 
The query and Ad pairs are collected from a commercial sponsored search engine, with relevance labels provided by experts. 
For evaluation, we treat the task as a binary classification problem and report AUC-ROC metrics, following production practices. 
The dataset consists of 13 million examples in the training set and 2 million examples in the test set, with data collected in multiple languages from global markets. This benchmark is particularly challenging because, while all tasks are from the ad domain, they span different product scenarios. Effectively modeling the correlation between tasks is crucial for achieving optimal performance.

We follow the same experimental design used for fine-tuning in the GLUE benchmark. For further details on the Ads dataset and experimental settings, please refer to Section B and Section A.1 of the supplementary material.

The results are presented in Table~\ref{tab:com-ads}. In the large-scale Ads dataset, LoRA-MT consistently outperforms LoRA-ST, further underscoring the effectiveness of multi-task fine-tuning. In this context, MTL-LoRA either surpasses or matches other methods on all 14 tasks. Notably, while other multi-task adaptations exhibit a seesaw effect when compared to the LoRA-ST approach—improving performance on some tasks but underperforming on others—MTL-LoRA consistently outperforms LoRA-ST across all tasks. This indicates that MTL-LoRA effectively mitigates interference between tasks during adaptation. Furthermore, compared to the second-ranking method in each task, MTL-LoRA achieved statistical significance with confidence 90\% in 10 of 14 tasks on the MPT model and 7 of 14 tasks on the LLaMA2 model. This significant performance advantage of MTL-LoRA in the Ads dataset validates its enhanced capability for MTL.

\subsection{Ablation Study}
We conduct ablation studies on MTL-LoRA to assess the impact of three key components: (1) the task-specific learnable transformation matrix $\Lambda_t$, (2) the temperature coefficient $\tau$ in Eq.~\ref{eq:mtl}, and (3) multiple low-rank up-projection matrices $n$.

For each ablation study, we use LLaMA2-7B as the backbone model and train it on GLUE dataset. In each ablation experiment, we systematically remove or disable one of the key components while keeping all other settings unchanged and report the GLUE metrics. The performance of each modified setting is compared against the full MTL-LoRA configuration. Additionally, we add LoRA's results on multi-task fine-tuning as a reference. Detailed hyperparameter settings are provided in Section A.3 of the supplementary material.

The results, presented in Table~\ref{tab:ablation}, clearly illustrate the importance of each component of MTL-LoRA. Omitting any one of these components results in a decline in performance. Notably, the inclusion of multiple low-rank up-projection matrices has the most significant impact. When $n=1$, the performance of MTL-LoRA falls below that of the LoRA-based multi-task fine-tuning on the GLUE benchmark. This suggests that effectively extracting task-specific information using $\Lambda_t$ relies on methods that aggregate diverse combinations of information across tasks.

\begin{figure*}[t]
  \centering
  \includegraphics[width=\textwidth]{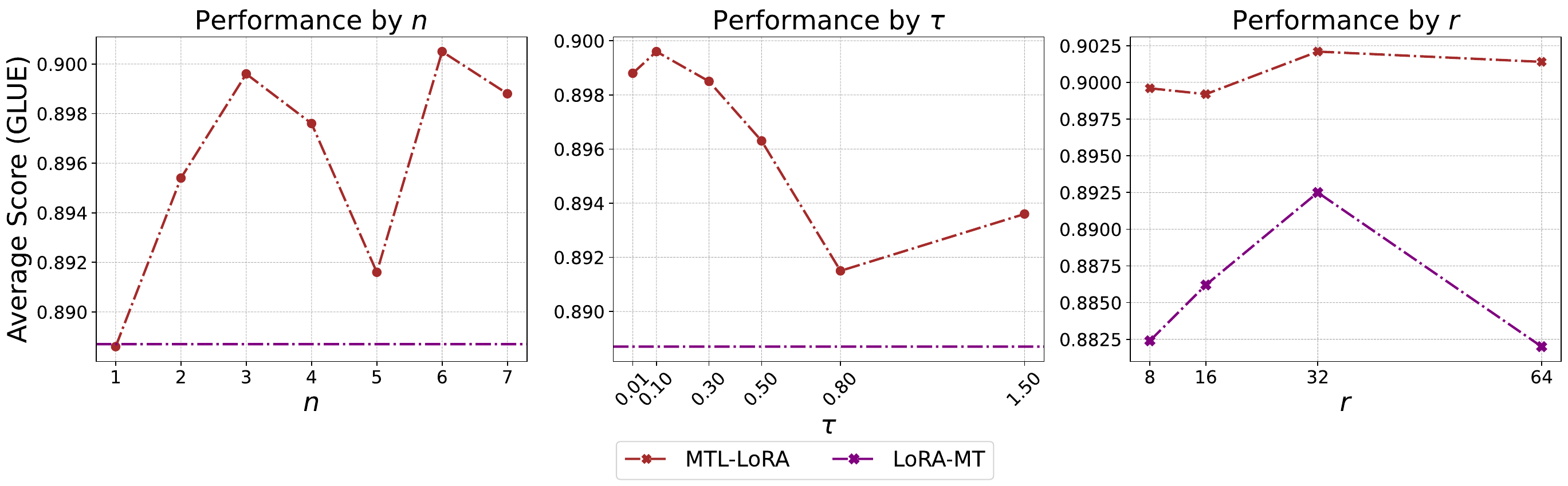}
   \vspace{-0.5cm}
  \caption{The performance of MTL-LoRA on GLUE benchmark with different hyperparameter configurations.}
  \label{fig:sensitivity}
\end{figure*}

\subsection{Sensitivity Analysis}
In this section, we analyze the robustness of MTL-LoRA under various parameter settings. Our investigation focus on how different values of $n$, $\tau$, and $r$ impact the performance of MTL-LoRA. We use LLaMA2-7B as the backbone model and conduct comparison on the GLUE benchmark. The results, presented in Figure~\ref{fig:sensitivity}, demonstrate that multiple up-projection metrics are crucial for MTL-LoRA, with $n=3$ yielding superior results in most scenarios. The value of temperature coefficient also affects the model performance, supporting that different up-projection matrices capture distinct aggregated information. Additionally, the analysis of different values for the rank $r$ reveals that MTL-LoRA maintains robustness at higher rank compared to vanilla LoRA.

\subsection{Task Differentiation}
The primary goal of MTL-LoRA is to enhance the effectiveness of LoRA in multi-task scenarios by preventing cross-task information interference while facilitating task information sharing. Consequently, we assert that the representation outputs of MTL-LoRA should be task-relevant.
To validate this, we use the outputs from MTL-LoRA as features for SVM classification, with labels corresponding to their respective tasks.
We use the Ads dataset for comparison because it encompasses a larger number of tasks, all within the same Ad domain, thus providing a challenging context for differentiating tasks. Specifically, we randomly sample 1,000 examples for each task from the Ads dataset and use the outputs of different LoRA adapters from the final block of the underlying LLMs as input features. An SVM classifier is trained on 40\% of the samples for each method, with the remaining 60\% reserved for evaluation. 
For detailed information on the experimental settings and hyperparameters of the SVM, please refer to Section A.4 of the supplementary material.

The results, as shown in Table~\ref{tab:svm}, indicate that MTL-LoRA significantly outperforms LoRA in multi-task fine-tuning, achieving a substantial improvement margin with both MPT and LLaMA2. The t-SNE visualization in Figure~\ref{fig:leading} further supports this observation, revealing that, in the multi-task adaptation scenario, LoRA tends to blend features from different tasks in the low-rank space, whereas MTL-LoRA effectively separates these tasks.  
When projected back into the full-rank space, MTL-LoRA forms distinct ‘task groups’, indicating that information is shared within each group while remaining distinct between groups. 
These findings confirm MTL-LoRA’s enhanced ability to distinguish between tasks and effectively reduce task interference. 
Furthermore, the relatively lower performance of LoRA in this setting highlights its limitations in extracting task-specific information.

\begin{figure}[t!]
\centering
\includegraphics[width=\linewidth]{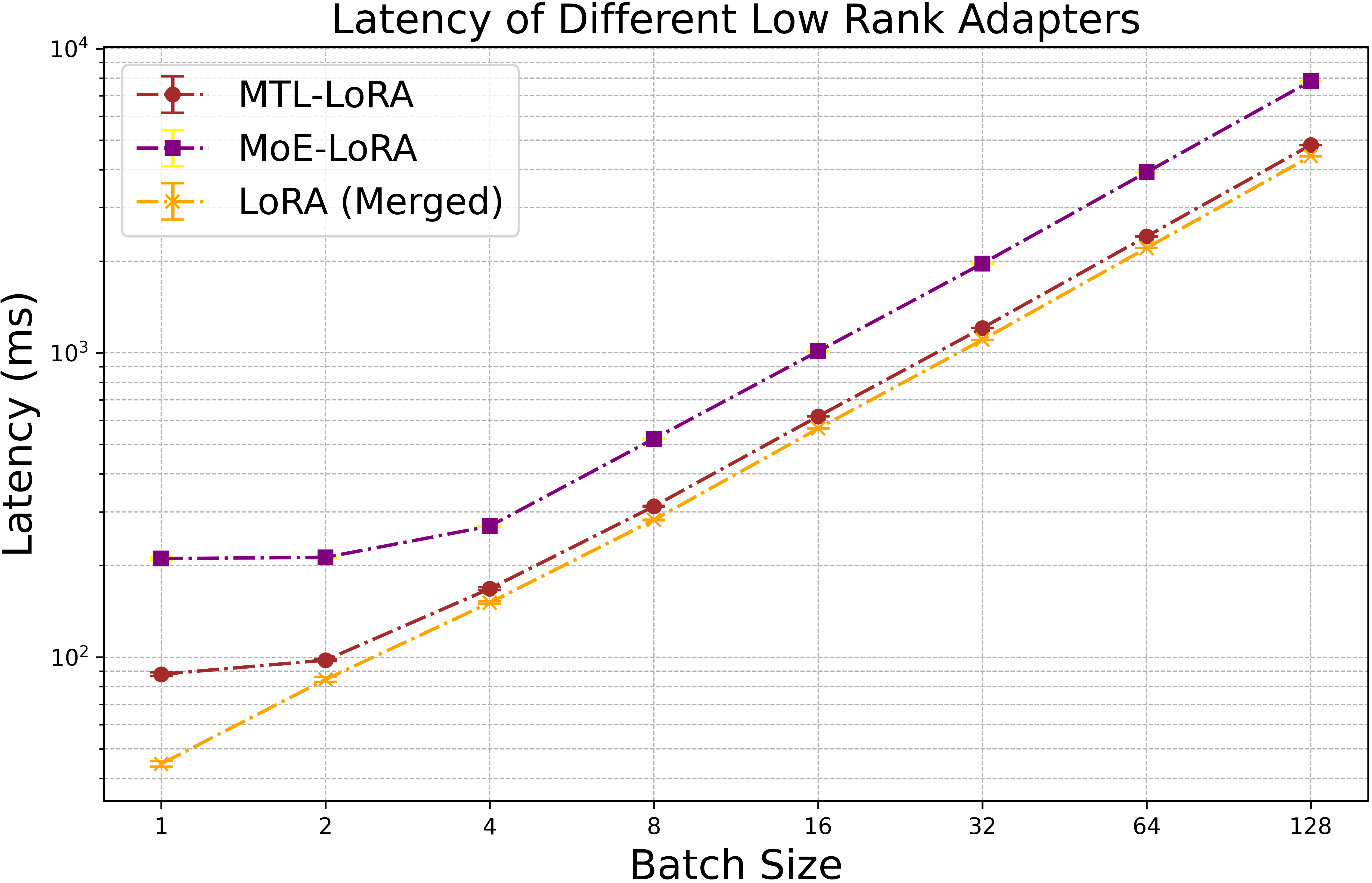}
\caption{Inference latency of different low-rank adapters on LLaMA2-7B with varying batch sizes. The results are averaged over 100 runs on a single A100-80G GPU. DoRA and MultiLoRA exhibit similar performance to LoRA as they can also be merged into pre-trained weights.}
\label{fig:latency}
\vspace{-5mm}
\end{figure}

\subsection{Inference Overhead}
Although MTL-LoRA demonstrates exceptional performance in multi-task learning, it relies on task-specific information to determine the appropriate transformation matrix for routing ($\Lambda_t$ in Eq.\ref{eq:mtl}). This design choice prevents the merging of all parameters into the original weights as LoRA does, which introduces additional inference latency.
However, thanks to MTL-LoRA’s task-level routing design, all operations are executed using matrix multiplication, avoiding the expert-level looping required by MoELoRA. This allows MTL-LoRA to fully utilize computational resources without incurring significant inference latency.

To validate this, we compared the inference latency of LoRA, MoELoRA, and MTL-LoRA across different batch sizes on the commonsense reasoning task, as depicted in Figure\ref{fig:latency}. The results clearly demonstrate that MTL-LoRA significantly outperforms MoELoRA in terms of inference speed. Compared to merged LoRA, MTL-LoRA introduces only minimal additional latency during inference, with this difference diminishes further under computationally intensive conditions with larger batch sizes. We believe that in latency-sensitive applications, the gap in inference speed between MTL-LoRA and merged LoRA can be further reduced through system-level optimizations, which we plan to explore in future work.

\label{sec:experiment}
\section{Conclusion}
We propose MTL-LoRA, a novel, advanced parameter-efficient fine-tuning method for multi-task low-rank adaptation. MTL-LoRA enhances the multi-task learning capability of LoRA by incorporating task-specific transformations in low-rank space along with a strategy for adaptively exploring multiple information sharing methods. This approach facilitates the learning of both task-specific and shared information.
Comprehensive experiments on multiple public academic benchmarks and a large-scale text Ads relevance dataset demonstrate that MTL-LoRA outperforms LoRA and its variants, including MultiLoRA, MoELoRA, and DoRA, validating its effectiveness in multi-task learning. Furthermore, extensive analysis of intermediate low-rank features and visualizations support our design motivations.
For future studies , we will focus on optimizing the design of MTL-LoRA to reduce the additional inference time while maintaining its effectiveness.\label{sec:conclusion}

\bibliography{aaai25}

\begin{thebibliography}{34}
\providecommand{\natexlab}[1]{#1}

\bibitem[{Abdin et~al.(2024)Abdin, Jacobs, Awan, Aneja, Awadallah, Awadalla, Bach, Bahree, Bakhtiari, Behl, Benhaim, Bilenko, Bjorck, Bubeck, Cai, Mendes, Chen, Chaudhary, Chopra, Giorno, de~Rosa, Dixon, Eldan, Iter, Goswami, Gunasekar, Haider, Hao, Hewett, Huynh, Javaheripi, Jin, Kauffmann, Karampatziakis, Kim, Khademi, Kurilenko, Lee, Lee, Li, Liang, Liu, Lin, Lin, Madan, Mitra, Modi, Nguyen, Norick, Patra, Perez-Becker, Portet, Pryzant, Qin, Radmilac, Rosset, Roy, Saarikivi, Saied, Salim, Santacroce, Shah, Shang, Sharma, Song, Ruwase, Wang, Ward, Wang, Witte, Wyatt, Xu, Xu, Yadav, Yang, Yang, Yu, Zhang, Zhang, Zhang, Zhang, Zhang, Zhang, and Zhou}]{Abdin2024Phi3TR}
Abdin, M.; Jacobs, S.~A.; Awan, A.~A.; Aneja, J.; Awadallah, A.; Awadalla, H.~H.; Bach, N.; Bahree, A.; Bakhtiari, A.; Behl, H.~S.; Benhaim, A.; Bilenko, M.; Bjorck, J.; Bubeck, S.; Cai, M.; Mendes, C. C.~T.; Chen, W.; Chaudhary, V.; Chopra, P.; Giorno, A.~D.; de~Rosa, G.; Dixon, M.; Eldan, R.; Iter, D.; Goswami, A.; Gunasekar, S.; Haider, E.; Hao, J.; Hewett, R.~J.; Huynh, J.; Javaheripi, M.; Jin, X.; Kauffmann, P.; Karampatziakis, N.; Kim, D.; Khademi, M.; Kurilenko, L.; Lee, J.~R.; Lee, Y.~T.; Li, Y.; Liang, C.; Liu, W.; Lin, E.; Lin, Z.; Madan, P.; Mitra, A.; Modi, H.; Nguyen, A.; Norick, B.; Patra, B.; Perez-Becker, D.; Portet, T.; Pryzant, R.; Qin, H.; Radmilac, M.; Rosset, C.; Roy, S.; Saarikivi, O.; Saied, A.; Salim, A.; Santacroce, M.; Shah, S.; Shang, N.; Sharma, H.; Song, X.; Ruwase, O.; Wang, X.; Ward, R.; Wang, G.; Witte, P.; Wyatt, M.; Xu, C.; Xu, J.; Yadav, S.; Yang, F.; Yang, Z.; Yu, D.; Zhang, C.-Y.; Zhang, C.; Zhang, J.; Zhang, L.~L.; Zhang, Y.; Zhang, Y.; and Zhou, X. 2024.
\newblock Phi-3 Technical Report: A Highly Capable Language Model Locally on Your Phone.
\newblock \emph{ArXiv}, abs/2404.14219.

\bibitem[{Aghajanyan, Zettlemoyer, and Gupta(2020)}]{aghajanyan2020intrinsic}
Aghajanyan, A.; Zettlemoyer, L.; and Gupta, S. 2020.
\newblock Intrinsic dimensionality explains the effectiveness of language model fine-tuning.
\newblock \emph{arXiv preprint arXiv:2012.13255}.

\bibitem[{Biderman et~al.(2024)Biderman, Ortiz, Portes, Paul, Greengard, Jennings, King, Havens, Chiley, Frankle, Blakeney, and Cunningham}]{Biderman2024LoRALL}
Biderman, D.; Ortiz, J.~G.; Portes, J.; Paul, M.; Greengard, P.; Jennings, C.; King, D.; Havens, S.; Chiley, V.; Frankle, J.; Blakeney, C.; and Cunningham, J.~P. 2024.
\newblock LoRA Learns Less and Forgets Less.
\newblock \emph{ArXiv}, abs/2405.09673.

\bibitem[{Brown et~al.(2020)Brown, Mann, Ryder, Subbiah, Kaplan, Dhariwal, Neelakantan, Shyam, Sastry, Askell et~al.}]{brown2020language}
Brown, T.; Mann, B.; Ryder, N.; Subbiah, M.; Kaplan, J.~D.; Dhariwal, P.; Neelakantan, A.; Shyam, P.; Sastry, G.; Askell, A.; et~al. 2020.
\newblock Language models are few-shot learners.
\newblock \emph{Advances in Neural Information Processing Systems}, 33: 1877--1901.

\bibitem[{Crawshaw(2020)}]{crawshaw2020multi}
Crawshaw, M. 2020.
\newblock Multi-task learning with deep neural networks: A survey.
\newblock \emph{arXiv preprint arXiv:2009.09796}.

\bibitem[{Hayou, Ghosh, and Yu(2024)}]{Hayou2024LoRAEL}
Hayou, S.; Ghosh, N.; and Yu, B. 2024.
\newblock LoRA+: Efficient Low Rank Adaptation of Large Models.
\newblock \emph{ArXiv}, abs/2402.12354.

\bibitem[{He et~al.(2021)He, Zhou, Ma, Berg-Kirkpatrick, and Neubig}]{he2021towards}
He, J.; Zhou, C.; Ma, X.; Berg-Kirkpatrick, T.; and Neubig, G. 2021.
\newblock Towards a unified view of parameter-efficient transfer learning.
\newblock \emph{arXiv preprint arXiv:2110.04366}.

\bibitem[{He et~al.(2015)He, Zhang, Ren, and Sun}]{he2015delving}
He, K.; Zhang, X.; Ren, S.; and Sun, J. 2015.
\newblock Delving deep into rectifiers: Surpassing human-level performance on imagenet classification.
\newblock In \emph{Proceedings of the IEEE International Conference on Computer Vision}, 1026--1034.

\bibitem[{Hoffmann et~al.(2022)Hoffmann, Borgeaud, Mensch, Buchatskaya, Cai, Rutherford, de~Las~Casas, Hendricks, Welbl, Clark, Hennigan, Noland, Millican, van~den Driessche, Damoc, Guy, Osindero, Simonyan, Elsen, Rae, Vinyals, and Sifre}]{Hoffmann2022TrainingCL}
Hoffmann, J.; Borgeaud, S.; Mensch, A.; Buchatskaya, E.; Cai, T.; Rutherford, E.; de~Las~Casas, D.; Hendricks, L.~A.; Welbl, J.; Clark, A.; Hennigan, T.; Noland, E.; Millican, K.; van~den Driessche, G.; Damoc, B.; Guy, A.; Osindero, S.; Simonyan, K.; Elsen, E.; Rae, J.~W.; Vinyals, O.; and Sifre, L. 2022.
\newblock Training Compute-Optimal Large Language Models.
\newblock \emph{ArXiv}, abs/2203.15556.

\bibitem[{Hofmann, Sch{\"o}lkopf, and Smola(2008)}]{hofmann2008kernel}
Hofmann, T.; Sch{\"o}lkopf, B.; and Smola, A.~J. 2008.
\newblock Kernel methods in machine learning.

\bibitem[{Houlsby et~al.(2019)Houlsby, Giurgiu, Jastrzebski, Morrone, De~Laroussilhe, Gesmundo, Attariyan, and Gelly}]{houlsby2019parameter}
Houlsby, N.; Giurgiu, A.; Jastrzebski, S.; Morrone, B.; De~Laroussilhe, Q.; Gesmundo, A.; Attariyan, M.; and Gelly, S. 2019.
\newblock Parameter-efficient transfer learning for NLP.
\newblock In \emph{International Conference on Machine Learning}, 2790--2799. PMLR.

\bibitem[{Hu et~al.(2021)Hu, Shen, Wallis, Allen-Zhu, Li, Wang, Wang, and Chen}]{hu2021lora}
Hu, E.~J.; Shen, Y.; Wallis, P.; Allen-Zhu, Z.; Li, Y.; Wang, S.; Wang, L.; and Chen, W. 2021.
\newblock Lora: Low-rank adaptation of large language models.
\newblock \emph{arXiv preprint arXiv:2106.09685}.

\bibitem[{Hu et~al.(2023)Hu, Wang, Lan, Xu, Lim, Bing, Xu, Poria, and Lee}]{hu2023llm}
Hu, Z.; Wang, L.; Lan, Y.; Xu, W.; Lim, E.-P.; Bing, L.; Xu, X.; Poria, S.; and Lee, R. K.-W. 2023.
\newblock Llm-adapters: An adapter family for parameter-efficient fine-tuning of large language models.
\newblock \emph{arXiv preprint arXiv:2304.01933}.

\bibitem[{Huang et~al.(2023)Huang, Liu, Lin, Pang, Du, and Lin}]{Huang2023LoraHubEC}
Huang, C.; Liu, Q.; Lin, B.~Y.; Pang, T.; Du, C.; and Lin, M. 2023.
\newblock LoraHub: Efficient Cross-Task Generalization via Dynamic LoRA Composition.
\newblock \emph{ArXiv}, abs/2307.13269.

\bibitem[{Li and Liang(2021)}]{li2021prefix}
Li, X.~L.; and Liang, P. 2021.
\newblock Prefix-tuning: Optimizing continuous prompts for generation.
\newblock \emph{arXiv preprint arXiv:2101.00190}.

\bibitem[{Lin, Madotto, and Fung(2020)}]{lin2020exploring}
Lin, Z.; Madotto, A.; and Fung, P. 2020.
\newblock Exploring versatile generative language model via parameter-efficient transfer learning.
\newblock \emph{arXiv preprint arXiv:2004.03829}.

\bibitem[{Liu et~al.(2023{\natexlab{a}})Liu, Li, Wu, and Lee}]{Liu2023VisualIT}
Liu, H.; Li, C.; Wu, Q.; and Lee, Y.~J. 2023{\natexlab{a}}.
\newblock Visual Instruction Tuning.
\newblock \emph{ArXiv}, abs/2304.08485.

\bibitem[{Liu et~al.(2023{\natexlab{b}})Liu, Wu, Zhao, Zhu, Xu, Tian, and Zheng}]{Liu2023WhenMM}
Liu, Q.; Wu, X.; Zhao, X.; Zhu, Y.; Xu, D.; Tian, F.; and Zheng, Y. 2023{\natexlab{b}}.
\newblock When MOE Meets LLMs: Parameter Efficient Fine-tuning for Multi-task Medical Applications.
\newblock In \emph{Annual International ACM SIGIR Conference on Research and Development in Information Retrieval}.

\bibitem[{Liu et~al.(2024)Liu, Wang, Yin, Molchanov, Wang, Cheng, and Chen}]{liu2024dora}
Liu, S.-Y.; Wang, C.-Y.; Yin, H.; Molchanov, P.; Wang, Y.-C.~F.; Cheng, K.-T.; and Chen, M.-H. 2024.
\newblock Dora: Weight-decomposed low-rank adaptation.
\newblock \emph{arXiv preprint arXiv:2402.09353}.

\bibitem[{Liu et~al.(2019)Liu, He, Chen, and Gao}]{liu-etal-2019-multi}
Liu, X.; He, P.; Chen, W.; and Gao, J. 2019.
\newblock Multi-Task Deep Neural Networks for Natural Language Understanding.
\newblock In Korhonen, A.; Traum, D.; and M{\`a}rquez, L., eds., \emph{Proceedings of the 57th Annual Meeting of the Association for Computational Linguistics}, 4487--4496. Florence, Italy: Association for Computational Linguistics.

\bibitem[{Liu et~al.(2022)Liu, Ji, Fu, Tam, Du, Yang, and Tang}]{liu2022p}
Liu, X.; Ji, K.; Fu, Y.; Tam, W.; Du, Z.; Yang, Z.; and Tang, J. 2022.
\newblock P-tuning: Prompt tuning can be comparable to fine-tuning across scales and tasks.
\newblock In \emph{Proceedings of the 60th Annual Meeting of the Association for Computational Linguistics (Volume 2: Short Papers)}, 61--68.

\bibitem[{Min et~al.(2023)Min, Ross, Sulem, Veyseh, Nguyen, Sainz, Agirre, Heintz, and Roth}]{min2023recent}
Min, B.; Ross, H.; Sulem, E.; Veyseh, A. P.~B.; Nguyen, T.~H.; Sainz, O.; Agirre, E.; Heintz, I.; and Roth, D. 2023.
\newblock Recent advances in natural language processing via large pre-trained language models: A survey.
\newblock \emph{ACM Computing Surveys}, 56(2): 1--40.

\bibitem[{Pfeiffer et~al.(2020)Pfeiffer, Kamath, R{\"u}ckl{\'e}, Cho, and Gurevych}]{pfeiffer2020adapterfusion}
Pfeiffer, J.; Kamath, A.; R{\"u}ckl{\'e}, A.; Cho, K.; and Gurevych, I. 2020.
\newblock AdapterFusion: Non-destructive task composition for transfer learning.
\newblock \emph{arXiv preprint arXiv:2005.00247}.

\bibitem[{Qin et~al.(2023)Qin, Zhang, Zhang, Chen, Yasunaga, and Yang}]{Qin2023IsCA}
Qin, C.; Zhang, A.; Zhang, Z.; Chen, J.; Yasunaga, M.; and Yang, D. 2023.
\newblock Is ChatGPT a General-Purpose Natural Language Processing Task Solver?
\newblock \emph{ArXiv}, abs/2302.06476.

\bibitem[{Shi et~al.(2024)Shi, Huang, Song, Li, Zhang, Huang, Wei, Deng, Sun, and Zhang}]{Shi2024ResLoRAIR}
Shi, S.; Huang, S.; Song, M.; Li, Z.; Zhang, Z.; Huang, H.; Wei, F.; Deng, W.; Sun, F.; and Zhang, Q. 2024.
\newblock ResLoRA: Identity Residual Mapping in Low-Rank Adaption.
\newblock \emph{ArXiv}, abs/2402.18039.

\bibitem[{Team(2023)}]{MosaicML2023Introducing}
Team, M.~N. 2023.
\newblock Introducing MPT-7B: A New Standard for Open-Source, Commercially Usable LLMs.
\newblock Accessed: 2023-05-05.

\bibitem[{Touvron et~al.(2023)Touvron, Martin, Stone, Albert, Almahairi, Babaei, Bashlykov, Batra, Bhargava, Bhosale et~al.}]{touvron2023llama}
Touvron, H.; Martin, L.; Stone, K.; Albert, P.; Almahairi, A.; Babaei, Y.; Bashlykov, N.; Batra, S.; Bhargava, P.; Bhosale, S.; et~al. 2023.
\newblock Llama 2: Open foundation and fine-tuned chat models.
\newblock \emph{arXiv preprint arXiv:2307.09288}.

\bibitem[{Wang et~al.(2018)Wang, Singh, Michael, Hill, Levy, and Bowman}]{wang2018glue}
Wang, A.; Singh, A.; Michael, J.; Hill, F.; Levy, O.; and Bowman, S.~R. 2018.
\newblock {GLUE}: A Multi-Task Benchmark and Analysis Platform for Natural Language Understanding.
\newblock ArXiv preprint 1804.07461.

\bibitem[{Wang et~al.(2023)Wang, Lin, Zeng, and Zhang}]{wang2023multilora}
Wang, Y.; Lin, Y.; Zeng, X.; and Zhang, G. 2023.
\newblock MultiLoRA: Democratizing LoRA for Better Multi-Task Learning.
\newblock \emph{arXiv preprint arXiv:2311.11501}.

\bibitem[{Wei et~al.(2022{\natexlab{a}})Wei, Tay, Bommasani, Raffel, Zoph, Borgeaud, Yogatama, Bosma, Zhou, Metzler et~al.}]{wei2022emergent}
Wei, J.; Tay, Y.; Bommasani, R.; Raffel, C.; Zoph, B.; Borgeaud, S.; Yogatama, D.; Bosma, M.; Zhou, D.; Metzler, D.; et~al. 2022{\natexlab{a}}.
\newblock Emergent abilities of large language models.
\newblock \emph{arXiv preprint arXiv:2206.07682}.

\bibitem[{Wei et~al.(2022{\natexlab{b}})Wei, Wang, Schuurmans, Bosma, Xia, Chi, Le, Zhou et~al.}]{wei2022chain}
Wei, J.; Wang, X.; Schuurmans, D.; Bosma, M.; Xia, F.; Chi, E.; Le, Q.~V.; Zhou, D.; et~al. 2022{\natexlab{b}}.
\newblock Chain-of-thought prompting elicits reasoning in large language models.
\newblock \emph{Advances in Neural Information Processing Systems}, 35: 24824--24837.

\bibitem[{Yang et~al.(2024)Yang, Yang, Hui, Zheng, Yu, Zhou, Li, Li, Liu, Huang, Dong, Wei, Lin, Tang, Wang, Yang, Tu, Zhang, Ma, Xu, Zhou, Bai, He, Lin, Dang, Lu, Chen, Yang, Li, Xue, Ni, Zhang, Wang, Peng, Men, Gao, Lin, Wang, Bai, Tan, Zhu, Li, Liu, Ge, Deng, Zhou, Ren, Zhang, Wei, Ren, Fan, Yao, Zhang, Wan, Chu, Cui, Zhang, and Fan}]{Yang2024Qwen2TR}
Yang, A.; Yang, B.; Hui, B.; Zheng, B.; Yu, B.; Zhou, C.; Li, C.; Li, C.; Liu, D.; Huang, F.; Dong, G.; Wei, H.; Lin, H.; Tang, J.; Wang, J.; Yang, J.; Tu, J.; Zhang, J.; Ma, J.; Xu, J.; Zhou, J.; Bai, J.; He, J.; Lin, J.; Dang, K.; Lu, K.; Chen, K.-Y.; Yang, K.; Li, M.; Xue, M.; Ni, N.; Zhang, P.; Wang, P.; Peng, R.; Men, R.; Gao, R.; Lin, R.; Wang, S.; Bai, S.; Tan, S.; Zhu, T.; Li, T.; Liu, T.; Ge, W.; Deng, X.; Zhou, X.; Ren, X.; Zhang, X.; Wei, X.; Ren, X.; Fan, Y.; Yao, Y.; Zhang, Y.; Wan, Y.; Chu, Y.; Cui, Z.; Zhang, Z.; and Fan, Z.-W. 2024.
\newblock Qwen2 Technical Report.

\bibitem[{Zhang et~al.(2023)Zhang, Chen, Bukharin, Karampatziakis, He, Cheng, Chen, and Zhao}]{Zhang2023AdaLoRAAB}
Zhang, Q.; Chen, M.; Bukharin, A.; Karampatziakis, N.; He, P.; Cheng, Y.; Chen, W.; and Zhao, T. 2023.
\newblock AdaLoRA: Adaptive Budget Allocation for Parameter-Efficient Fine-Tuning.

\bibitem[{Zhao et~al.(2023)Zhao, Zhou, Li, Tang, Wang, Hou, Min, Zhang, Zhang, Dong et~al.}]{zhao2023survey}
Zhao, W.~X.; Zhou, K.; Li, J.; Tang, T.; Wang, X.; Hou, Y.; Min, Y.; Zhang, B.; Zhang, J.; Dong, Z.; et~al. 2023.
\newblock A survey of large language models.
\newblock \emph{arXiv preprint arXiv:2303.18223}.

\end{thebibliography}
\clearpage
\section{Supplementary Material}\label{sec:supp}
\section{A. Experimental Setting}
\subsection{A.1. Natural Language Understanding}
For all experiments conducted on the GLUE and Ads datasets, we employ the AdamW optimizer with weight decay set to 0 and utilize a linear learning rate decay scheduler. 
The batch size is set to 8 for the GLUE benchmark and 16 for the Ads dataset.
Unless otherwise specified, in all our experiments, we integrate adapter modules into every dense layer of the multi-head attention (namely $Q, K, V, O$) in the selected LLMs.
Details on other hyperparameters for each method are provided in below.

\paragraph{Full Tuning (FT):}
For Full Tuning on the GLUE benchmark, the learning rate is set to $8 \times 10^{-6}$ for both MPT and LLaMA2 models. For LLaMA2, we observed that a single epoch of training on each dataset led to highest performance. In contrast, for the MPT model, we extend the training to 5 epochs for CoLA and MRPC, 10 epochs for RTE, while maintaining a single epoch for the remaining datasets.

\paragraph{LoRA-ST:}
In all experiments, we set the rank $r$ to 8. 
For the Ads dataset, training is conducted for one epoch with a learning rate of $3\times 10^{-4}$ for both models. 
As for the GLUE benchmark, the learning rate is set to $4\times10^{-4}$. 
We train for 10 epochs on the RTE dataset and 5 epochs on the MRPC dataset, while all other datasets are trained for a single epoch.

\paragraph{LoRA-MT:}
In each case, we set the rank $r$ to 16 and the learning rate to $2\times10^{-4}$. 
All results are reported after training for 1 epoch.

\paragraph{Multi-LoRA:}
In all MultiLoRA experiments, we configure the number of LoRA modules to 3 and set the rank $r$ of each LoRA module to 8. 
The learning rate is set to $3\times 10^{-4}$ for Ads dataset and $2\times 10^{-4}$ for the GLUE benchmark.

\paragraph{MoE-LoRA:}
In all MoE-LoRA experiments, we set the number of experts in MoE-LoRA to 8, the rank of LoRA modules to 16, and the task embedding for each task to 64.
The learning rate is set to $3\times 10^{-4}$ for the Ads dataset and $2\times 10^{-4}$ for the GLUE benchmark.

\paragraph{MTL-LoRA:}
In all experiments involving MTL-LoRA, training is conducted for 1 epoch.
Unless otherwise specified, we set the low-rank $r$ to 8.
For the Ads dataset, the learning rate is $3\times 10^{-4}$ for LLaMA2 and $4\times 10^{-4}$ for MPT. The values of $n$ and $\tau$ are both set to 3 and 0.5, respectively, for the LLaMA2 and MPT models.
In the case of the GLUE benchmark, both models have their learning rate and $n$ set to $2\times 10^{-4}$ and $3$, respectively. 
The temperature hyperparameter $\tau$ is adjusted to 0.1 for LLaMA2 and 0.5 for MPT. 
For all experiments, unless otherwise specified, the low-rank parameter $r$ is set to 8.

\subsection{A.2. Commonsense Reasoning}
We follow the same instructional prompt and dataset configuration as DoRA~\cite{liu2024dora} for evaluating commonsense reasoning tasks. The results for DoRA and LoRA are taken from the original DoRA paper, while we have reimplemented MultiLoRA, MoELoRA, and MTL-LoRA. For each reimplementation, we applied the adapter to the $\mathbf{Q}, \mathbf{K}, \mathbf{V}, \mathbf{O}$ linear layers of the pre-trained model. Detailed hyperparameters are provided in Table~\ref{tab:hyper-commonsense}.

\begin{table}[t!]
\centering
\caption{Hyperparameter configurations on the commonsense reasoning tasks.}
\label{tab:hyper-commonsense}
\resizebox{0.85\columnwidth}{!}{%
\begin{tabular}{@{}lccc@{}}
\toprule
Hyperparameters             & MultiLoRA & MoELoRA & MTL-LoRA \\ \midrule
Rank $r$                    & 8         & 16       & 8        \\
Scale $\alpha$              & 16        & 32       & 16       \\
Learning Rate               & \multicolumn{3}{c}{3e-4}                                  \\
Warmup Ratio                & \multicolumn{3}{c}{0.03}                                  \\
Epochs                      & \multicolumn{3}{c}{3}                                     \\
Batch Size                  & \multicolumn{3}{c}{8}                                     \\
Weight Decay                & \multicolumn{3}{c}{0}                                     \\
Optimizer                   & \multicolumn{3}{c}{AdamW}                                 \\
$\beta$                     & \multicolumn{3}{c}{(0.9,0.95)}                            \\
Sequence Length             & \multicolumn{3}{c}{512}                                   \\
Expert Num                  & -         & 8         & -        \\
LoRA Num                    & 3         & -         & -        \\
Task Embedding Size         & -         & 64        & -        \\
Num of Up-Projection Matrix & -         & -         & 3        \\
Temperature                 & -         & -         & 0.8      \\ \bottomrule
\end{tabular}%
}
\end{table}

\subsection{A.3. Ablation Study}
In all ablation experiments, we use LLaMA2-7B as our base model. The default settings for the number of low-rank up-projection matrices $n$,  temperature coefficients $\tau$, and  learning rates are 3, 0.1, $2 \times 10^{-4}$, respectively.
All experimental results are obtained after training for one epoch.

\subsection{A.4. Task Differentiation}
We randomly sample 1,000 samples for each task from the Ads dataset, utilizing the output of different LoRA adapters from the final block of the underlying LLMs as input features for the SVM classifier. 
The dataset is split into training and test sets in a 4:6 ratio, with the SVM's parameter C set to 1. 
We refrain from using any kernel functions and employ the 'one-vs-rest' strategy for multi-class classification. For comparison, we conduct identical classification procedures using the LoRA-MT method.

\begin{table}[t!]
\centering
\caption{Ads dataset statistics (M for Million).}
\label{tab:ads-data}
\vspace{-2mm}
\resizebox{\columnwidth}{!}{%
\begin{tabular}{@{}lcccccccccccccc@{}}
\toprule
Task & 0  & 1 & 2 & 3 & 4 & 5 & 6 & 7 & 8 & 9 & 10 & 11 & 12 & 13  \\ \cmidrule(r){1-15}
\#Train & 1.09M & 0.52M & 0.86M & 2.12M & 2.11M & 2.12M & 1.33M & 0.42M & 0.87M & 0.62M & 0.18M & 0.42M & 0.42M & 0.19M \\
\#Test & 0.06M & 0.12M & 0.04M& 0.10M & 0.10M & 0.10M& 0.19M& 0.17M& 0.09M& 0.34M& 0.25M& 0.17M& 0.17M& 0.18M \\
\#Label & 2 & 2 & 2 & 5 & 5 & 4 & 4 & 2 & 4 & 2 & 2& 2 & 2& 2 \\
\bottomrule 
\end{tabular}%
}
\end{table}

\section{B. Details for Ads Dataset}
The text Ads relevance dataset used in this paper is a large-scale, real-world dataset from online advertising platform. 
This dataset comprises 14 tasks and covers various scenarios. It includes a total of 13 million examples in the training set and 2 million examples in the test set. 
We use 10\% of the training data for hyper-parameter tuning. Each dataset is uniformly formatted as \{query, Ad, label\}, where query is the search phrase combined with related features that reflect search intent and Ad is the concatenation of Ad description, features, or corresponding landing page, etc. Among these tasks, task 0 to 7 are sampled from the North American market, while the rest are from the international markets. The labels, provided by human experts, indicate how relevant is the user intent to the given Ad. For the training data, the granularity of relevance may vary, including 2, 4, 5 degrees. For the test data, we follow the practice and treat label 0 as negative and other labels as positive.
Statistical details of the Ads dataset are provided in Table~\ref{tab:ads-data}, while examples are presented in Table~\ref{tab:ads_sample}.

\section{C. Task Template}
We employ the same prompt templates for adapting different PEFT methods. 
The prompt templates utilized for the GLUE benchmark are detailed in Table~\ref{tab:template_glue}.
Regarding the Ads dataset, we evaluate all the templates listed in Table~\ref{tab:template_ads} and find that Template 6 yields the best results. 
Consequently, Template 6 are used for all tasks in the Ads dataset.

\begin{table}[ht!]
    \centering
    \begin{minipage}{0.99\columnwidth}
    \vspace{0mm}    
    \centering
    \caption{Prompt templates for the GLUE benchmark.}
    \vspace{-5mm}
    \label{tab:template_glue}
    \begin{tcolorbox} 
        \centering
        \fontsize{8pt}{10pt}\selectfont
        \hspace{-6mm}
        \begin{tabular}{p{0.99\columnwidth}}
                \VarSty{CoLA:}

                Is the following sentence "\{\}" grammatically acceptable? Answer:

                \VarSty{SST2:}

                Is the following sentence "\{\}" sentimently positive? Answer: 

                \VarSty{MRPC:}

                Dose the following sentence "\{\}" convey the equivalent meaning as "\{\}"? Answer:

                \VarSty{STSB:}

                On a scale of 0 to 5, how similar the sentence "\{\}" with the sentence "\{\}"? Answer:

                \VarSty{QQP:}
                
                Is the following question "\{\}" essentially asking the same thing as "\{\}"? Answer:

                \VarSty{MNLI:}

                Dose the statement "\{\}" imply that "\{\}" ? Answer:

                \VarSty{QNLI:}

                Based on the statement: "\{\}" dose the following sentence "\{\}" have a definitive answer? Answer:

                \VarSty{RTE:}

                Dose the text "\{\}" entail the statement "\{\}"? Answer:
        \end{tabular}
    \end{tcolorbox}
    \vspace{-3mm}
    \end{minipage}
\end{table}

\begin{table}[ht]
    \centering
    \begin{minipage}{0.99\columnwidth}
    \vspace{0mm}    
    \centering
    \caption{Prompt templates for the Ads dataset.}
    \vspace{-5mm}
    \label{tab:template_ads}
    \begin{tcolorbox} 
        \centering
        \fontsize{8pt}{10pt}\selectfont
        \hspace{-6mm}
        \begin{tabular}{p{0.99\columnwidth}}
                \VarSty{Template 1:}

                Does the query "\{\}" match the Ad "\{\}"? Answer:

                \VarSty{Template 2:}

                Is the following query "\{\}" semantically relevant to the Ad "\{\}"? Answer: 

                \VarSty{Template 3:}

                Dose the following sentence "\{\}" convey the equivalent meaning as "\{\}"? Answer:

                \VarSty{Template 4:}

                Query: "\{\}".
                
                Based on this query, can we conclude the Ad "\{\}" is matched? Answer:

                \VarSty{Template 5:}

                Answer using yes or no: does the query "\{\}" match the Ad "\{\}"? Answer: 
                
                \VarSty{Template 6:}

                Dose the query "\{\}" match the Ad "\{\}" ? Answer:
        \end{tabular}
    \end{tcolorbox}
    \vspace{-3mm}
    \end{minipage}
\end{table}

\begin{table*}[t!]
    \centering
    \begin{minipage}{0.99\textwidth}
    \vspace{0mm}    
    \centering
    \caption{Selected Examples from the Ads dataset.}
    \vspace{-5mm}
    \label{tab:ads_sample}
    \begin{tcolorbox} 
        \centering
        \fontsize{8pt}{10pt}\selectfont
        \hspace{-6mm}
        \begin{tabular}{p{0.99\textwidth}}
                



                \VarSty{Example 1:}
                
                Query:
                lymphoma lymphoma Lymphoma | CDC - Centers for Disease Control and Prevention Lymphoma: Causes, Symptoms, Types, Treatments, and Prognosis - Healthline Lymphoma Cancer | Understanding Lymphoma lymphoma lymphoma what is lymphoma lymphoma lymphoma cancer lymphoma

                Ad:
                kerendia 10mg price

                Label:
                Positive

                \VarSty{Example 2:}
                
                Query:
                building online store ; e commerce retail software ;  ;  ;  ; online store builder ; e commerce retail software ; Find An Online Store Builder | Side-by-Side Comparison 2020 ; create an online store ; e commerce retail software ; Mozello - Create your online store quickly and easily ; build online store website ; e commerce retail software ; Online Store Builder—Create One Quickly (2023) - Shopify

                Ad:
                start online store ; printful.com ; e commerce retail software ; How to Start an Online Store in 2020 (Step by Step) ; How to Start Your Own Online Store (Beginner Guide) ;  ; starting an online shop ; e commerce ; How To Start An Online Store In 8 Steps (2023 Guide) ; create online store ; e commerce retail software ; Mozello - Create your online store quickly and easily ; store online ; computers consumer electronics ; Google Shopping - Shop Online, Compare Prices  Where to Buy

                Label:
                Negative

                \VarSty{Example 3:}

                Query:
                koeksflaekt ; Fjäråskupan Köksfläkt - Svenskt hantverk och kvalitet ; fjaraskupan.se/Köksfläkt ; Köp en köksfläkt från Fjäråskupan. Finns i ett flertal mått och kan lätt anpassas.. Klassiskt rostfritt, stilren mässing eller något annat? Valet är ditt. Se utbud online

                Ad:
                krav kanal köksfläkt enfamiljshus

                Label:
                Positive

                \VarSty{Example 4:}
                
                Query:
                binarycent ; spread betting ; Binarycent - Forex and CFD broker with possibility to trade in cents!

                Ad:
                scamrecovery.net/Chargeback/BinaryCent; | BinaryCent Review | Is BinaryCent.com a Legit Broker Or a Scam?; BinaryCent Review;Have You Been Scammed by BinaryCent?;  Home >> Online Trading Scams >> Albert Street Victoria, Mahe, Seychelles Company: Cent Projects Ltd Regulation: Not Regulated BinaryCent Rating BinaryCent Warning France AMF warns against BinaryCent! Get Help Now ! Read More

                Label:
                Fine

                \VarSty{Example 5:}

                Query:
                things to see in milan

                Ad:
                milan what to do

                Label: 
                Positive

                \VarSty{Example 6:}

                Query: gacha club download

                Ad: softonic ; Download Gacha Club-free-latest version ; softonic.com ; The world's largest software, App discovery destination. Your trusted website!. The Best downloads for any device. New apps. Free Download. Software download. Game

                Label: 
                Positive

                \VarSty{Example 7:}

                Query:
                gartendeko für draußen

                Ad:
                gartendekorationen ;Entdecke Gartendekorationen - Große Artikelauswahl bei OTTO - Lass dich von uns inspirieren ;www.otto.de/Top-Artikel ;Gartendekorationen online gefunden auf OTTO.de. Entdecke die Vielfalt und bestelle jetzt!. Für dich ausgewählte Artikel - Aktuelle Trends Top-Marken zu fairen Preisen. ;Gartendekorationen online gefunden auf OTTO.de. Entdecke die Vielfalt und bestelle jetzt!. Für dich ausgewählte Artikel - Aktuelle TrendsTop-Marken zu fairen Preisen.

                Label:
                Positive
        \end{tabular}
    \end{tcolorbox}
    \vspace{-3mm}
    \end{minipage}
\end{table*}

\end{document}